\documentclass[12pt]{article}

\usepackage{graphicx}
\usepackage{multirow}
\usepackage{amsmath,amssymb,amsfonts}
\usepackage{amsthm}
\usepackage{mathrsfs}
\usepackage[title]{appendix}
\usepackage{xcolor}
\usepackage{textcomp}
\usepackage{booktabs}
\usepackage{algorithm}
\usepackage{algorithmicx}
\usepackage{algpseudocode}
\usepackage{listings}
\usepackage{mdframed}
\usepackage{url}
\usepackage[margin=1in]{geometry}
\usepackage[colorlinks=true, linkcolor=green!50!black, citecolor=green!50!black, urlcolor=green!50!black]{hyperref}

\theoremstyle{definition}

\begin{document}

\title{Barriers to Complexity-Theoretic Proofs that ``AGI" Using Machine Learning is Impossible}

\author{Michael Guerzhoy\\
University of Toronto\\
\texttt{guerzhoy@cs.toronto.edu}}

\date{}

\maketitle

\begin{abstract}

    A recent paper~\cite{van2024reclaiming} claims to have proved that achieving human-like intelligence using learning from data is intractable in a complexity-theoretic sense. We point out that the proof relies on an unjustified assumption about the distribution of (input, output) tuples in the data. We briefly discuss that assumption in the context of two fundamental barriers to repairing the proof: the need to precisely define ``human-like," and the need to account for the fact that a particular machine learning system will have particular inductive biases that are key to the analysis. Another attempt to repair the proof, by focusing on subsets of the data, faces barriers in terms of defining the subsets.

\end{abstract}

\section{Introduction}

In~\cite{van2024reclaiming} a claim is made that the authors \textbf{``formally prove [in the paper that] creating systems with human(-like or -level) cognition (``AGI" for short, for the purposes of this paper) is intrinsically computationally intractable."}

Here, we show that the paper falls short of formally proving the claim.

In this paper, first, we point out that the proof in~\cite{van2024reclaiming}, if it were sound, could equally well be used to prove that learning to classify images distributed as in the ImageNet dataset is intractable. Since that task is tractable in practice~\cite{NIPS2012_c399862d}, it must be the case that combination of the claim, the formalization, and the proof is flawed.

Second, we identify the source of the error in the proof: an unjustified assumption about the distribution $\mathcal{D}$ of (situation, behaviour) tuples in the data. Specifically, we identify a key unproven premise that underlies the proof: that the distribution $\mathcal{D}$ of tuples $(s, b)$, with $s$ denoting ``situation" and $b$ denoting ``[human] behaviour" in response to $s$ can be an arbitrary polytime-sampleable distribution.

If $\mathcal{D}$ is a model of human behaviour, both the marginal distribution of $s$ and the conditional distribution $P_\mathcal{D}(b|s)$ are in fact highly structured. For example, if $s$ encodes natural images, the marginal distribution $P_\mathcal{D}(s)$ would need to account for the hierarchical structure of natural images~\cite{simoncelli2001natural}. If $P_\mathcal{D}(b|s)$ models human chess moves, the distribution would need to account (among other things) for the rules of chess. This means that many $\mathcal{D}$'s can be ruled out a-priori.

As we argue below, the fact that, in the authors' proof, $\mathcal{D}$ denotes both the distribution of situation-behaviour tuples and an arbitrary polytime-sampleable distribution means that the authors did not prove what they set out to prove.

We argue that two critical issues must be resolved when attempting to repair the proof (although we make no claim that the proof could be repaired).

\begin{itemize}
    \item ``Human-like" behaviour must be defined precisely.

    \item The fact that while an arbitrary function is not learnable, some structured functions can be learned with appropriate inductive biases must be considered.

\end{itemize}

Another attempt to repair the proof by focusing on subsets of the data also faces a barrier.

The paper is organized as follows. In Section~\ref{sec:ingenia} we introduce the ``Ingenia Theorem" of~\cite{van2024reclaiming}, along with some necessary context. In Section~\ref{sec:reductio}, we illustrate that not having met one of the challenges leaves the current proof vulnerable to a \textit{reductio ad absurdum} argument. In Section~\ref{sec:fault} we point out what we believe to be a central flaw in the proof. In Section~\ref{barriers}, we identify what we see as the challenges that a correct version of the proof would have to overcome.  In Section~\ref{sec:worst-case}, we discuss why the worst-case analysis of~\cite{van2024reclaiming}, as performed on the domain of polytime-sampleable distributions, does not imply that learning human cognition is intractable. In Section~\ref{sec:lower-bound}, we explain that it follows from Section~\ref{sec:worst-case} that~\cite{van2024reclaiming} did not derive a lower-bound on the real-world complexity of constructing human-like AI from human data.

\section{The Ingenia Theorem \label{sec:ingenia}}

We first present the ``AI-by-Learning" problem, and then the ``Ingenia Theorem" that the authors derive

\subsubsection*{AI-by-Learning Problem (from~\cite{van2024reclaiming})}

\textbf{Given}: An integer $K$ and a way of sampling from a distribution $\mathcal{D}$.

\noindent \textbf{Task}: Find a description $L_A \in \mathcal{L}_A$, with length $|L_A| \leq K$, of an algorithm $A \in \mathcal{A}$ that with probability $\geq \delta(n)$, taken over the randomness in the sampling, satisfies:

\[
\Pr_{s \sim \mathcal{D}_n} \left[ A(s) \in B_s \right] \geq \frac{|B_s|}{|B|} + \epsilon(n).
\]

Here $\delta(n)$ and $\epsilon(n)$ are arbitrary non-negligible functions. A function $f$ is non-negligible if there is some $d$ such that for sufficiently large $n$, $f(n) \geq 1/n^d$.

On p. 6 (bottom), $\mathcal{D}$ is described as follows: ``A dataset $D$ drawn from $\mathcal{D}$ consists of situation-behaviour tuples, a.k.a `samples`: $(s_1, b_1), (s_2, b_2), ..., (s_{|D|}, b_{|D|})$".

\newpage

\subsubsection*{Ingenia Theorem (from~\cite{van2024reclaiming})}
\begin{mdframed}
AI-by-Learning is  intractable
\end{mdframed}

Informally, \cite{van2024reclaiming} claim that the Ingenia Theorem implies that it is not possible to obtain a human-like AI by learning from examples, by reducing the Perfect-vs-Chance problem, known to be intractable~\cite{hirahara2022np}, to an instance of the AI-by-Learning problem.

\subsubsection*{Perfect-vs-Chance (decision problem) (from~\cite{van2024reclaiming})}
\begin{mdframed}
\textbf{Given:} A way to sample a given distribution $\mathcal{D}$ over $\{0, 1\}^n \times \{0, 1\}$, an integer $k$, and the promise that one of the following two cases apply:
\begin{enumerate}
    \item There is an efficient program $M$ of size at most $k$ such that $\Pr_{(x,y)\sim \mathcal{D}} [M(x) = y] = 1$
    \item For any program $M$ of size at most $k$,\\
    $\Pr_{(x,y)\sim \mathcal{D}} [M(x) = y] \leq \frac{1}{2} + \frac{1}{2^n}$\\
    where $0 < \delta < 1$ is an arbitrary constant.
\end{enumerate}
\textbf{Question:} Is (1) or (2) the case?
\end{mdframed}

Crucially, as we will see later, the reduction of Perfect-vs-Chance to AI-by-Learning relies on the distribution $\mathcal{D}$ in AI-by-Learning being an arbitrary polytime-sampleable distribution, rather than a distribution of situation-behaviour tuples.

\section{The Proof ``Proves" ImageNet is Intractable \label{sec:reductio}}

In the proofs in the paper, ``AGI" could be replaced with ``image recognition in ImageNet" without altering the mathematical structure of the proofs, implying that learning image classification on ImageNet is intractable. However, image classification on ImageNet does work~\cite{NIPS2012_c399862d}. The reason this substitution would not alter the proofs is that the distribution $\mathcal{D}$ of situation-behaviour tuples is not characterized mathematically except to specify that it is polytime-sampleable.

We note that image recognition is not, in general, a solved problem: the point here, rather, is that \textit{some} image recognition problems are solvable. Any claim that a particular problem is intractable (the problem can be ``learning to classify ImageNet from data" or ``learning to predict human behaviour from responses from data") must refer to the specific structure of the problem.

This \textit{reductio ad absurdum} is a  way to see that at least one of the following must be true:
\begin{enumerate}
    \item The formalization in the paper does not capture the  meaning of ``X is intractable"
    \item The proof is flawed
    \item Image classification in ImageNet was achieved despite being intractable
\end{enumerate}

Note that while (3) is a remote theoretical possibility in the case of ImageNet image classification, the argument would work equally well for tasks that are certainly tractable, such as learning to classify images whose average intensity is below 128 (on a scale of 0 to 255) vs. above 128. ImageNet classification was chosen here for illustrative purposes, as a problem that has been solved, initially by using strong appropriately-selected inductive biases -- those implied by the Convolutional Neural Network architecture~\cite{NIPS2012_c399862d}.

This observation points to the fact that \cite{van2024reclaiming} did not provide a correct proof for the claim that ``creating systems with human(-like or -level) cognition (``AGI" for short, for the purposes of this paper) is intrinsically computationally intractable."

The proof involves a claim, a formalization, and a reduction. Below, we identify what we believe to be the central flaw in the proof.

\section{Incorrect Reduction from Perfect-vs-Chance \\to AI-by-Learning \label{sec:fault}}

The key issue in the paper is whether the distribution $\mathcal{D}$ in the AI-by-Learning problem is an arbitrary polytime-sampleable distribution or a distribution $(s, b)\sim \mathcal{D}$, the distribution of situation-behaviour tuples. If the distribution is (polytime-sampleable) arbitrary, the ``AI-by-Learning" problem is misnamed: it is the general problem of learning a tractable function from examples. On the other hand, if $\mathcal{D}$ is a distribution of situation-behaviour tuples, as described on p. 6 (bottom), then the problem is arguably appropriately-named, but it was not demonstrated that Perfect-vs-Chance reduces to the problem.

In the informal presentation on p. 6, the authors assume a learning machine $\mathcal{M}$ that is able to learn from examples to approximately map $s$ to $b$ when $(s, b)\sim \mathcal{D}$, the distribution of situation-behaviour tuples. They use this learning machine on instances of Perfect-vs-Chance and a different distribution $\mathcal{D}$. However, it can happen that the learning machine $\mathcal{M}$ works on data samples from $\mathcal{D}$, the distribution of situations-behaviours, but not on the identically-named arbitrary polytime-sampleable distribution $\mathcal{D}$ in the ``formalized" version of AI-by-Learning and in the appendix.

The proof would work if the reduction were to a version of AI-by-Learning where the distribution $\mathcal{D}$ is (polytime-sampleable) arbitrary rather than a distribution of situation-human behaviour tuples. However, if that were the case, the intractability of AI-by-Learning would only imply that there are tractable functions that are not learnable. As a stylized fact, this is well-known~\cite{valiant1984theory}.

In the Appendix of \cite{van2024reclaiming}, it is stated that AI-by-Learning will have a mechanism $\mathcal{M}$ that will learn ``regardless of the distribution $\mathcal{D}$". If the definition of AI-by-Learning is as in the Appendix and the formalized version, the proof can go through; however, AI-by-Learning in that case is ``Ability to approximately learn an arbitrary function from examples," a much stronger construct than AI-by-Learning that doesn't correspond to the concept introduced in the main paper.

\section{Fundamental Barriers to Intractability Results for AGI \label{barriers}}

It is sufficient to show that the reduction does not work to point out that the $\mathcal{D}$ on p. 22 is arbitrary but denotes a particular structured distribution in the context of the informal AI-by-Learning problem. This issue is not easily fixed.

However, we believe that it is conceptually possible to try to reduce known hard problems to the version AI-by-Learning that works on genuine situation-behaviour tuples (though it would of course end up not being actually possible if AI-by-Learning is not NP-hard). We see two main challenges to that project that were not addressed in the paper.

\subsection{Challenge 1: Mathematically characterizing the distribution $\mathcal{D}$ in AI-by-Learning}

The authors define the distribution $\mathcal{D}$ in the context of AI-by-Learning informally as the distribution of situation-behaviour tuples one would observe in humans. A reduction from a mathematical problem to a problem involving the distribution $\mathcal{D}$ could not be proven mathematically without a rigorous mathematical definition of $\mathcal{D}$, although it could perhaps be checked empirically.

\subsubsection{Is the structure of the AI-by-Learning $\mathcal{D}$ ``almost" arbitrary?}
One might counter that the structure of the distribution $\mathcal{D}$ in AI-by-Learning $\mathcal{D}$ is ``almost" arbitrary, and attempt to repair the proof that way. However, far from being arbitrary, the AI-by-Learning $\mathcal{D}$ is highly structured, as we argued above. A proof that $\mathcal{D}$ is ``almost" arbitrary in an interesting way would, in our opinion, be highly interesting, but it is not present in the paper. The paper, as well as the first author\footnote{\url{https://x.com/IrisVanRooij/status/1691134292225675266}}, imply that it is a significant fact that when $\mathcal{D}$ is treated as arbitrary in the paper, it is treated as an arbitrary polytime-sampleable distribution. As pointed out above, however, it is clear that polytime-sampleability is not a sufficient constraint to exclude all tuples that are not genuine situation-behaviour tuples; for example, the constraint does not exclude probability distributions that do not make syntactically invalid strings that are unlikely to be produced by humans unlikely.

Concrete examples of polytime-sampleable but unlearnable distributions that were not shown to naturally arise in human situation-behaviour tuples include distributions where the behaviour is generated by a non-invertible mechanism, such as a cryptographic hash function~\cite{Aaronson_2013}.

\subsection{Challenge 2: Are there subsets of situation-behaviour tuple data that are not learnable?}

One might attempt to repair the proof by claiming that there are subsets of the set of situation-behaviour tuple data where the behaviour is generated by a non-invertible mechanism.

For example, one might argue that humans are able to execute arbitrary (or near-arbitrary) algorithms, and a subset of the situation-behaviour tuples would consist of input-output tuples for arbitrary Turing Machines.

There are a number of challenges with this approach to repairing the proof. The key issue is that  it is far from obvious that humans can execute an \textit{arbitrary} algorithm in their minds.

\begin{enumerate}
    \item Humans' working memory is limited. This means that humans would often use pen and paper to execute algorithms. If the intermediate steps are included in the data, the learning problem may become easier. If the intermediate steps are excluded, the learning problem may become unnatural: for example, it is obvious that it is not possible to learn to predict the output of a human's using a one-time pad~\cite{Aaronson_2013} if the pad is secret and excluded from the training data, but this is not a natural or interesting problem. In general, it is obvious that if one subsets the training data adversarially, learning would not be possible. One might make distinctions between the extreme example of learning to predict the output of a one-time pad encryption, but that argument needs to be made.

    \item It is not obvious that humans execute arbitrary algorithms. It might be that situation-behaviour tuples only contain input-output tuples for algorithms that are either known are close to known algorithms. If that is the case, complete training data could be helpful for the learning problem. In the extreme case, if there exists Python code for any algorithm humans execute and that code is in the training data, the learning problem is very likely tractable (It should be acknowledged that it is \textit{not} currently the case that there is Python code that would generate human-like situation-behaviour tuples in general, and it has not been shown that writing such code is possible; however, known examples of humans executing \textit{algorithms we \textbf{know} to be complex enough to be non-invertible} likely do correspond to existing Python code).
\end{enumerate}

\subsubsection{Repairing the proof}
To repair the proof by addressing this challenge, one must argue that the non-learnable subsets of the data are ``interesting": that the learning problem there is of interest in itself.

\subsection{Challenge 3: Inductive Biases}
The No Free Lunch (NFL) Theorems in machine learning~\cite{adam2019no} imply that, on an arbitrary problem, no model is necessarily better than another. However, for any particular (computable) function, a good model exists.

A proof of the intractability of AI-by-Learning would likely need to contend with the fact that in practice, any learning algorithm would have a particular inductive bias, which may be particularly suitable to solving AI-by-Learning, even if a ``blind" search would be intractable.

For example, it is believed that Convolutional Neural Networks (ConvNets) are particularly easy to train on natural image data because their inductive bias is conducive to working well on natural image data~\cite{NEURIPS2023_eb1bad7a}. Since no AGI trained by AI-by-Learning currently exists (nor does AGI obtained in some other way exist), it is possible that we do not know and will never know of good inductive biases for training AGI. However, that is not, to our knowledge, proven to be mathematically impossible.

In~\cite{guerzhoy2024occam}, we argue that, contra~\cite{bender2020climbing}, there is evidence from the history of physics that is relevant to the question of whether, with appropriate inductive biases, a learning machine could learn to predict physical processes. To the extent that induction using automatic learning is possible, it is a contingent fact about the universe~\cite{hume1894enquiry}. However, to argue that induction is not possible in our particular universe, one needs to explicitly bring in evidence about our universe.

Note that accounting for inductive biases is a challenge, but inductive biases are not directly an issue in the proofs in~\cite{van2024reclaiming}

\section{Discussion: Worst-Case Analysis \label{sec:worst-case}}

The proof can be interpreted as switching to a ``worst-case analysis'' from the initial problem of analyzing the tractability of learning a specific distribution. It is important to note what the worst-case analysis proves in this case.

To do that, we reiterate the prominent claim in the abstract and a related but distinct problem for which what is effectively the worst-case analysis is provided.

The prominent claim is: \textbf{``formally prove [in the paper that] creating systems with human(-like or -level) cognition (``AGI" for short, for the purposes of this paper) is intrinsically computationally intractable."}

The problem for which the worst-case analysis is provided is ``Learn to generalize from samples from a polytime-sampleable distribution $\mathcal{D}$". The worst-case analysis shows that for any learning procedure, there exists a distribution $\mathcal{D}$, which is not proven to do anything with human situation-behaviour tuples, that is not tractably learnable.

In order to prove the authors' claim, however, one must show that the distribution $\mathcal{D}$ of human situation-behaviour tuples is not tractably learnable. That is a different claim.

This is analogous to claiming that a specific SAT instance is intractable to solve because SAT is NP-complete, without showing that the specific instance is indeed hard (for examples, SAT instances that happen to be 2-SAT instances are tractable~\cite{Aaronson_2013} because of their specific structure). We emphasize that it was \text{not} shown that the structure of the distribution $\mathcal{D}$ of human situation-behaviour tuples is such that learning from it is tractable. But equally, the authors did not show that the structure of $\mathcal{D}$ of human situation-behaviour tuples is such that learning from it is intractable.

\section{Discussion: Do \cite{van2024reclaiming} derive a lower-bound on the real-world complexity of
constructing human-like AI from human data \label{sec:lower-bound}}

In the paper, the authors claim~\cite{van2024reclaiming}:
\begin{quote}
"By granting Dr. Ingenia these highly idealised conditions that
simplify and abstract away from real-world complications, we can
formally derive a reliable \textbf{lower-bound on the real-world complexity of
constructing human-like AI from human data}. To do so, we need to make
the problem AI-by-Learning formally precise such that it becomes
amenable to computational complexity analysis [...]
\end{quote}

As we have shown above, the task the authors ended up proving is intractable includes learning from an arbitrary polytime-computable distribution, which was not proven to correspond to the distribution of human situation-behaviour tuples. They have also not proven that learning from an arbitrary computable distribution is never harder than learning from human data. Although the authors simplified the problem faced by Ingenia in some ways, they did not provably not make it harder in other ways. Therefore, the result is not a real bound.
\section{Conclusion}

The abstract of~\cite{van2024reclaiming} claims to have proved that AGI through learning is intractable. In this paper, we have argued that this has not been proven. We point out the specific flaw in the proof: while the abstract and the informal introduction refer to a distribution of situation-behaviour tuples $\mathcal{D}$, the proof in the Appendix relies on $\mathcal{D}$'s being an arbitrary polytime-sampleable distribution.

We have also not proven the negation of the claim in the paper: as the authors of~\cite{van2024reclaiming} correctly note, there is no proof that AGI is ``inevitable."

However, we believe there are fundamental barriers to this style of proof, as outlined in Section~\ref{barriers}.

\bibliography{sample}

\end{document}